\documentclass[letterpaper, 10 pt, conference]{ieeeconf}

\IEEEoverridecommandlockouts 
\usepackage{times}
\usepackage{epsfig}
\usepackage{graphicx}
\usepackage{multicol}
\usepackage{xcolor}
\usepackage{listings}
\usepackage{url}
\urldef{\mailsa}\path|{thomas.buchmann, bernhard.westfechtel}@uni-bayreuth.de|

\newcommand{\code}[1]{\textsf{\small #1}}

\definecolor{dkblue}{rgb}{0.16,0,1}
\definecolor{dkgreen}{rgb}{0.24,0.5,0.37}
\definecolor{gray}{rgb}{0.3,0.3,0.3}
\definecolor{mauve}{rgb}{0.3,0,0.4}
\definecolor{eclipse}{rgb}{0.58,0,0.33}

\lstdefinelanguage{ATL}{
 morekeywords={lazy, rule, from, to},
 sensitive=true,
 morecomment=[l]{//},
 morecomment=[n]{/*}{*/},
 morestring=[b]",
 morestring=[b]',
 morestring=[b]"""
}

\lstdefinelanguage{QVTr}{
 morekeywords={transformation, top, relation, checkonly, enforce, domain, when, where},
 sensitive=true,
 morecomment=[l]{//},
 morecomment=[n]{/*}{*/},
 morestring=[b]",
 morestring=[b]',
 morestring=[b]"""
}

\lstdefinelanguage{ocl20} {morekeywords={and,bodycontext,context,def,derive,else,endif,endpackage,if,implies,in,
  init,inv,let,not,or,package,post,pre,static,then,xor},
 sensitive=false,
 morecomment=[l]{//},
 morecomment=[s]{/*}{*/},
 morestring=[b]'
}
 
\renewcommand*{\code}[1]{\texttt{#1}}

\title{\LARGE \bf Towards A Domain-specific Language For Pick-And-Place Applications}

\author{Thomas Buchmann$^{1}$, Johannes Baumgartl$^{2}$, Dominik Henrich$^{2}$ and Bernhard Westfechtel$^{1}$
\thanks{$^{1}$T. Buchmann and B. Westfechtel are with Chair of Applied Computer Science I (Software Engineering), 
        University of Bayreuth, 95447 Bayreuth, Germany
        {\tt\small firstname.lastname@uni-bayreuth.de}}%
\thanks{$^{2}$J.Baumgartl and D. Henrich are with Chair of Applied Computer Science III (Robotics and Embedded Systems), 
        University of Bayreuth, 95447 Bayreuth, Germany
        {\tt\small firstname.lastname@uni-bayreuth.de}}%
}

\begin{document}

\maketitle
\thispagestyle{empty}
\pagestyle{empty}

\begin{abstract}
Programming robots is a complicated and time-consuming task. A robot is essentially a real-time, distributed embedded system. Often, control and communication paths within the system are tightly coupled to the actual physical configuration of the robot. Thus, programming a robot is a very challenging task for domain experts who do not have a dedicated background in robotics. In this paper we present an approach towards a domain specific language, which is intended to reduce the efforts and the complexity which is required when developing robotic applications. Furthermore we apply a software product line approach to realize a configurable code generator which produces C++ code which can either be run on real robots or on a robot simulator.
\end{abstract}



\section{Introduction}
\label{sec:intro}

A robot is essentially a real-time, distributed embedded system. Robot systems consist of different hardware components and different sensors which results in a very complex and highly variable system architecture. Often, control and communication paths within the system are tightly coupled to the actual physical configuration of the robot. As a consequence, these robots can be assembled, configured, and programmed only by experts. While this is the state of the art for robot programming nowadays, it is evident that using model-driven software engineering, and domain specific languages in particular, could provide great benefits to this domain by raising the level of abstraction and reducing complex and recurring programming tasks.

\emph{Model-driven software engineering} \cite{Frankel2003,Voelter2006}
puts strong emphasis on the development of high-level models rather
than on the source code. Models are neither considered as documentation nor
as informal guidelines how to program the actual system. In contrast,
models have a well-defined syntax and semantics. Moreover,
model-driven software engineering aims at the development of
\emph{executable} models. \emph{Code generators} are used in model-driven software engineering, to transform the specification of higher-level models into source code. A \emph{domain-specific language (DSL)} is a programming or specification language which is dedicated to a particular problem domain. 

\emph{Software product line
  engineering} (SPLE) \cite{Clements2001,Pohl2005,Weiss1999} deals with the
systematic development of products belonging to a common system
family. Rather than developing each instance of a product line from
scratch, reusable software artifacts are created such that each
product may be composed from a library of components. Furthermore, it provides means to capture and manage the variability of a particular application domain. In common approaches, \emph{feature models} \cite{Kang1990} are used for that purpose.  

In this paper, we present the work in progress of our domain-specific language for pick-and-place applications and especially the configurable code generator which produces C++ code. 



\section{A Domain Specific Language for Pick-And-Place Applications}
\label{sec:dsl}

As stated in Section \ref{sec:intro}, programming a robot is a very complex task. Resulting programs highly depend on the robot's hardware and the environment in which the robot is being operated. Thus, our approach - whose basic ideas are presented in \cite{ICSOFT-Baumgartl2013} - aims at enabling programmers without dedicated knowledge in the robotics domain to specify robot applications. 

\begin{figure}[htb!]
	\centering
	\includegraphics[width=\columnwidth]{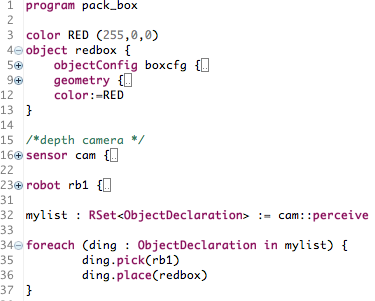}
	\caption{A small example of our DSL code.}
	\label{fig:dsl}
\end{figure}

The core part of our approach is a declarative domain-specific language for pick and place applications. We chose to start with this domain, since it covers basic robotic tasks like moving robots, grasping objects and placing them at a different location. These tasks, which sound easy at first glance, include inherently complex subtasks like object modeling, path planning, grasp planning, and placement planning. To empower users without dedicated background in those tasks, we abstracted from those concepts. Instead, modeling and planning operations are implemented in a C++ framework, which is used by the code generator presented in Section \ref{sec:codegen}. 
As a consequence, the DSL code can be kept simple and human readable as shown in Figure \ref{fig:dsl}. 

\subsection{Design Decisions}
\label{sec:dd}

The most difficult task when designing a domain-specific language is to find the right level of abstraction as well as the required keywords. A basic question is whether object and hardware declarations are required in the DSL or not. 

The sample DSL code in Figure \ref{fig:dsl} contains various declarations of different types. In its current status the DSL allows declarations, for e.g. colors, objects, sensors, and robots as well as object and robot configurations (c.f. lines 2 - 41 in the sample). 

While declarations of sensors and robots are useful when the generated code is meant to be run using several robots, declarations might be obsolete when using an educated distribution planner to assign a subtask to a specific sensor or robot. However, in this paper we focus on one robot with one sensor network capable to model objects to manipulate with, where those hardware declarations are just convenient. 


Dependent on used hardware, requirements for the algorithms might vary. Those dependencies should be implemented as constraints on the feature model of the product line and not be part of the DSL, since the concrete algorithms are transparent to the user, likewise the interaction with the concrete hardware.


The second design decision is concerned about the keywords that should be provided by the DSL. The current version of the DSL comprises keywords for object, sensor, and robot declarations and configurations. Furthermore keywords for built-in data types and control structures are included.
The keyword \code{object} in the declarative part can be used to define static environment or known objects.
Following an object-centered approach, objects are linked with hardware by keywords for object manipulation (\code{pick} and \code{place}), robot movement (\code{move}), and operations on sensors (e.g. \code{perceive}). A concrete (planning) algorithm must be available for each of the keywords. However, different realizations concerning one keyword might exist. Those are selected depending on the hardware.


\subsection{Implementation}
\label{sec:implementation}

We decided to use the Xtext\footnote{http://www.eclipse.org/xtext} framework for our textual DSL. Xtext allows the specification of a context-free grammar of a language. It uses ANTLR\footnote{http://www.antlr.org} as a parser generator, which means that is able to parse LL(*) grammars. Furthermore, the Xtext framework allows to enrich the Xtext grammar specification with context-sensitive information, which is used to unparse a text into an Ecore-based tree representation. The resulting, automatically generated editor comprises full-fledged support for syntax highlighting and code completion.


\section{Configurable Code Generator}
\label{sec:codegen}

The DSL code needs to be compiled into executable code in order to be run on real robots or within a simulator. To this end, we use the Accelo\footnote{http://www.eclipse.org/acceleo} framework which provides an implementation of the \emph{OMG MOF Model to Text standard} \cite{OMG2008}. Acceleo can be used to specify code generators for arbitrary Ecore-based metamodels. 

The target platform of the code generator is GeNBot - a C++ framework which comprises different algorithms for path planning, grasp planning and placement planning as well as a modular interface to different robots (Kuka LWR, Kuka KR16-2, St\"aubli RX130) and simulators. 

Acceleo provides a template-based code generation engine equipped with its own template language MTL. OCL\footnote{Object Constraint Language} is used to retrieve model information dynamically, which is used in the templates to generate code. 

\begin{figure}[b]
			\centering
			\includegraphics[width=\columnwidth]{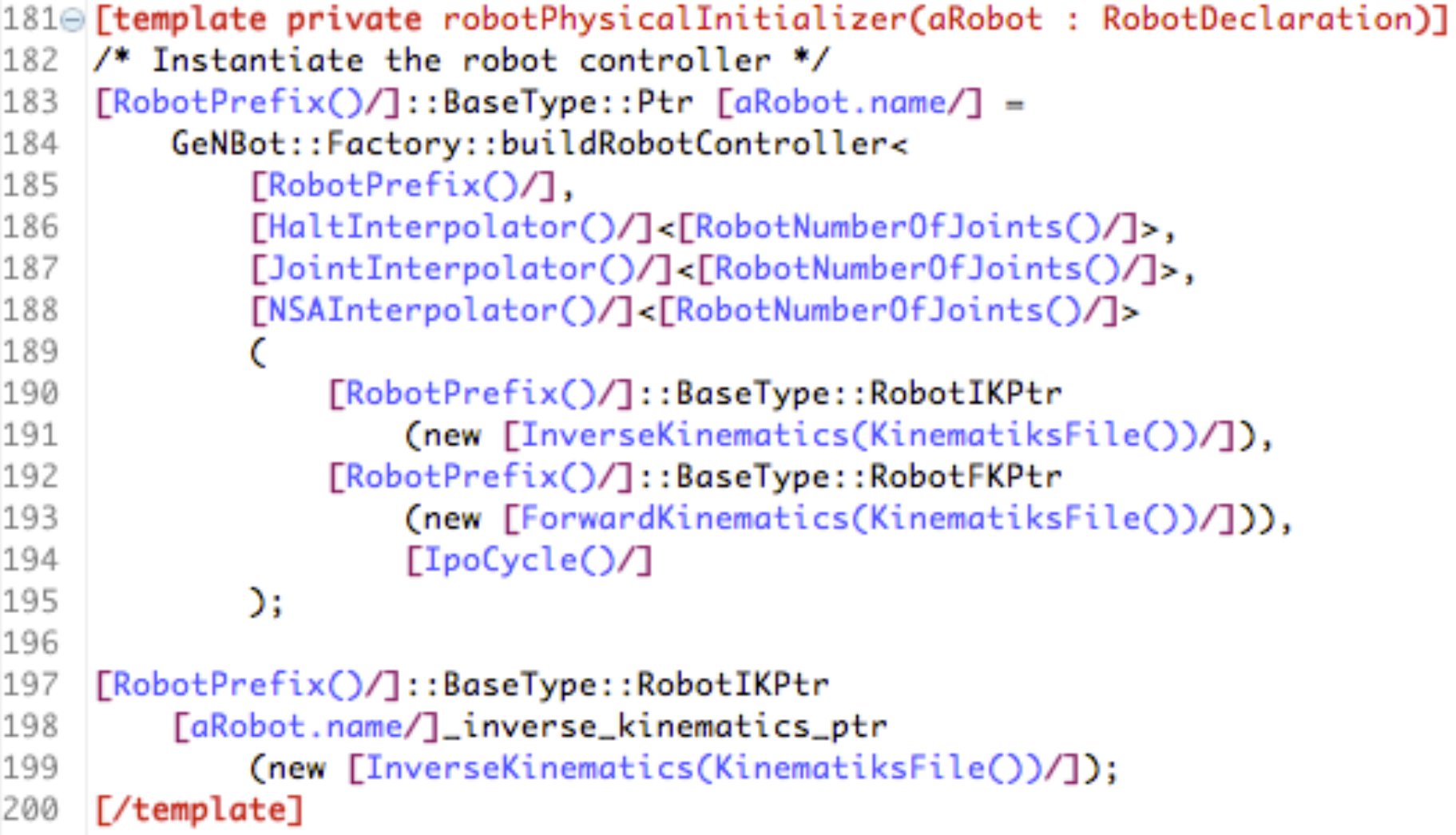}
			\caption{A cutout of the code generator templates.}
			\label{fig:codegen}
	    \end{figure}

Figure \ref{fig:codegen} shows a small cutout of our code generation templates which is used to initialize a LWR robot controller. The code formatted in blue color between square brackets depicts dynamic code fragments which are extracted from the model (e.g. the DSL code) at runtime. Text formatted in black color is static text which is used as it stands each time the template is invoked. 

	    \begin{figure}[htb!]
			\begin{lstlisting}[caption={Cutout of the generated C++ code}, language=C++, label=lst:code]
			/* Instantiate the robot controller */
			GeNBot::LWRRobotController::BaseType::Ptr rb1 = GeNBot::Factory::buildRobotController<
					GeNBot::LWRRobotController,
					GeNBot::HaltInterpolator<7>,
					GeNBot::ReflexxesJointInterpolator<7>,
					GeNBot::ReflexxesNSAInterpolator6D<7>
					(	
						GeNBot::LWRRobotController::BaseType::RobotIKPtr
							(new GeNBot::LWR_ik_AC()),
					   	GeNBot::LWRRobotController::BaseType::RobotFKPtr
					   		(new GeNBot::LWRFK(std::string("/path/to/kinematicsFile.xml"))),
					   	0.034f 
				   );
				   	
			GeNBot::LWRRobotController::BaseType::RobotIKPtr 
				inverse_kinematics_ptr
					(new GeNBot::LWR_ik_AC());
			\end{lstlisting}
		\end{figure}

The code which is produced by the template above is shown in Listing \ref{lst:code}. The code contains fragments which are neccessary to initialize a Kuka LWR robot controller in the GeNBot framework. This code is necessary in every application which is intended to be run on this type of hardware. But it also contains some variable parts like the number of joints for example so that it could not be reused by plain copy and paste in ``traditional'' programming approaches.

As stated in \cite{ICSOFT-Baumgartl2013}, our approach aims at integrating a product line approach to cover the variability which may occur in the target domain due to changing hardware (robots, sensors) and software (algorithms used for planning tasks etc.). Thus, we started to integrate our 
FAMILE environment, which is dedicated to the model-driven development of software product lines \cite{ECMFA-Buchmann2012,FOSD-Buchmann2012}.

\begin{figure}
	\centering
	\includegraphics[scale=0.5]{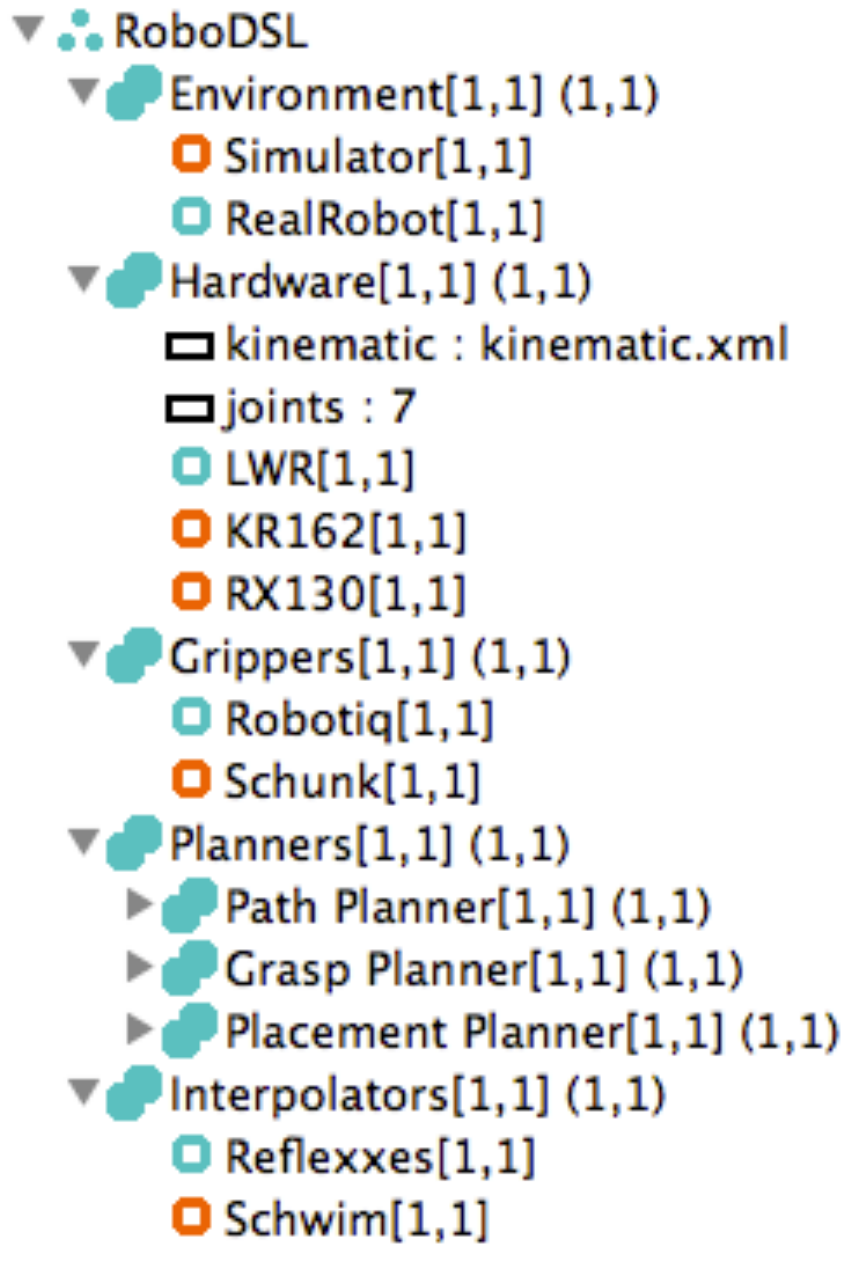}
	\caption{A sample feature configuration.}
	\label{fig:featureConf}
\end{figure}

Currently, we are addressing the variability which concerns the code generator. Depending on the target platform (simulator, real robot) different building blocks of the C++ framework are used in the generated C++ code. Furthermore, three types of robots (Kuka LWR, Kuka KR16-2, St\"aubli RX130) and different planning algorithms are supported. Our FAMILE toolchain uses feature models \cite{Kang1990} to capture commonalities and variabilities of the product line. Figure \ref{fig:featureConf} shows a sample feature configuration of the product line, which is used as an input during the code generation process to bind the variability. Elements marked with cyan colored circles depict features which are included in the current feature configuration, while orange colored circles mark excluded ones. Features also may have attribute values, e.g. feature \code{Hardware} contains the attribute \code{joints}. 
In its current state, the configuration (e.g. selecting / deselecting) the appropriate features in the feature model has to be done manually. 

\section{Future Work}
\label{sec:future}

In its current state, our approach already covers variability on the level of the code generator. E.g. different code is being generated depending on the feature configuration which is passed to it. But variability in robotics hardware does not only affect the generated code (by initializing and using appropriate building blocks of the GeNBot framework), it may also concern the language itself. The presence / absence of hardware or software might result in the inclusion or exclusion of language constructs. In this case, the end user cannot specify DSL programs which cannot be run on the target hardware. As a consequence, variability on the level of the grammar of the DSL is required. Our toolchain FAMILE was built to support model-driven software product line engineering for arbitrary Ecore-based domain models. As an Xtext grammar is parsed into an Ecore-based syntax tree (as the Xtext editor is specified with Xtext itself) FAMILE can be used to map features from the feature model to grammar production rules. As a result, language elements decorated with feature expressions evaluating to false (in case the respective features are excluded from the current feature configuration) are omitted. 

Furthermore, Acceleo also provides an Ecore-based model for its abstract and concrete syntax. While the connection between feature model and code generation templates is realized via Acceleo queries at the moment, we are currently working on using the feature mapping capabilities of FAMILE for it as well. Unfortunately it does not work out of the box, as Acceleo (contrastingly to Xtext) does not use a parser which automatically creates instances of this Ecore model in memory. 

In order to support automatic configuration depending on the used hardware, a dedicated interface to the hardware as well as a protocol providing the required information (which will be used for an automatic configuration of the feature model) is necessary. This will also be addressed in future work. 

Finally, we plan to extend the language to address other robotic application domains apart from pick and place as well. 





\section{Related work}
\label{sec:related}

In \cite{ICSOFT-Baumgartl2013}, we present the basic ideas behind our approach, which is intended to provide textual DSLs for robotic applications, which can be adapted at runtime according to the actual robot configuration.
While \cite{ICSOFT-Baumgartl2013} offers a conceptual overview, we present the first version of a DSL for pick-and-place applications and a configurable code generator which creates C++ code in this paper. 

In \cite{Ingles2012}, the authors present an approach which uses a DSL to handle run-time variability in programs for service robots. The approach presented by Ingl\'{e}s-Romero et al. aims to support developers of robotic systems (e.g. experts in the robotics domain) while our approach is not restricted to robotic experts only. Even regular programmers without a dedicated background in robotics are able to write robotic programs with our DSL. Furthermore, the DSL is only able to express variability information. It is not possible to specify the behavior of the robot. 

Steck et al. present an approach \cite{Steck2009} that is dedicated to a model-driven development process of robotic systems. They present an environment called \emph{SmartSoft} \cite{Steck2010} which provides a component based approach to develop robotics software. The SmartSoft environment is based on Eclipse and the Eclipse Modeling Project\footnote{http://www.eclipse.org/modeling/}. It uses Papyrus\footnote{http://www.eclipse.org/papyrus} for UML modeling. By using a model-driven approach, the authors focus on a strict separation of roles throughout the whole development life-cycle. Again, experts in the robotics domain are addressed with this approach while our approach doesn't require expert knowledge in robotics.

\emph{RobotML} \cite{Dhouib2012}, a modeling language for robot programs also aims to provide model-driven engineering capabilities for the domain of robot programming. RobotML is an extension to the Eclipse-based UML modeling tool Papyrus. Papyrus puts strong emphasis on UML's profile mechanism, which allows domain-specific adaptations. RobotML  provides code generators for different target platforms, like Orocos, RTMaps, Arrocam or Blender/Morse. The approach presented by Dhouib et al. addresses developers of robot programs or algorithms, while our approach can also be used by regular programmers (of course robotic experts can use it as well and may gain an increase in productivity).

Bubeck et al. present in \cite{Bubeck2012} an overview about best practices for system integration and distributed software development in service robotics. 
Furthermore, the authors develop \emph{BRIDE}\footnote{http://ros.org/wiki/bride}, a graphical DSL for ROS developers. Using BRIDE, new ROS nodes or ROS-based systems can be specified in a graphical way and corresponding C++ or Python code may be generated. In addition, the required launch files for the ROS environment including the relevant parameters and dependencies are generated as well, similar to the approach which we used in our case study as described in \cite{ICSOFT-Baumgartl2013}. Similar to the approaches discussed above, BRIDE also addresses robot experts only. 

In \cite{Schultz2007}, Schultz et al. present an approach for a domain-specific language intended for programming self-configurable robots. The DSL is targeted towards the ATRON self-reconfigurable robot. Like all other approaches mentioned in this section, it aims to provide a higher-level of abstraction for robot experts.

In his PhD thesis \cite{gherardi2013variability} Gherardi presents an approach for variability modeling and resolution in component-based robotics systems. It differs from our approach in terms of the different layers of abstraction and also meta-layers where variability is addressed. Furthermore, the toolchain we use for software product line development follows an established development process. Finally, the DSL described in this paper addresses programmers without a dedicated background in robotics, while \cite{gherardi2013variability} requires a robotics expert to provide algorithms and the variability model and additionally a software engineering expert. 



\section{Conclusion}
\label{sec:conclusion}

In this paper we presented our approach towards easy robot programming for personal robots. We demonstrated the feasibility of our approach by presenting a small and declarative domain-specific language for pick and place applications. Furthermore, a product line approach was used to realize a configurable code generator for C++. 



\vspace{6ex}

\bibliographystyle{IEEEtran}
\bibliography{references,newreferences,abbrev}

\end{document}